\documentclass[runningheads]{llncs}

\usepackage{eccv}

\usepackage{eccvabbrv}
\usepackage[colorlinks,
            linkcolor = teal,
            urlcolor  = teal, 
            citecolor = teal,
]{hyperref}
\usepackage{graphicx}
\usepackage{booktabs}
\usepackage{multirow}
\usepackage[table]{xcolor}
\definecolor{lightgraymetric}{gray}{0.68}

\usepackage[accsupp]{axessibility}  
\usepackage{caption}

\usepackage{hyperref}

\usepackage{orcidlink}

\begin{document}

\title{HUGE-Bench: A Benchmark for High-Level UAV Vision-Language-Action Tasks} 

\titlerunning{HUGE-Bench}

\author{
Jingyu Guo\inst{1,2}\orcidlink{0009-0002-8645-4929}$^*$,
Ziye Chen\inst{1,2}$^*$,
Ziwen Li\inst{3},
Zhengqing Gao\inst{3},
Jiaxin Huang\inst{3}\orcidlink{0000-0003-1893-7662},\\
Hanlue Zhang\inst{3}\orcidlink{0009-0002-7062-4997},
Fengming Huang\inst{2},
Yu Yao\inst{4}\orcidlink{0000-0001-9797-364X},
Tongliang Liu\inst{4,3}\orcidlink{0000-0002-9640-6472}$^\dagger$,\\
Mingming Gong\inst{1,3}\orcidlink{0000-0001-7147-5589}$^\dagger$
}

\authorrunning{J.~Guo et al.}

\institute{
$^1$The University of Melbourne, Australia \\
$^2$Melsy Tech, China \\
$^3$MBZUAI, United Arab Emirates \\
$^4$The University of Sydney, Australia \\[0.5em]
$^*$Equal contribution;
$^\dagger$Corresponding authors. \\
\email{guo27@student.unimelb.edu.au, tongliang.liu@sydney.edu.au, mingming.gong@unimelb.edu.au} \\
Work done during internship at Melsy Tech. \\
\url{https://jingyu198.github.io/HUGE_Bench}
}

\maketitle


\begin{abstract}
We present \textbf{HUGE-Bench}, a benchmark for High-Level UAV Vision--Language--Action (HL-VLA) tasks that tests whether an agent can interpret concise language and execute complex, process-oriented trajectories with safety awareness. HUGE-Bench comprises 4 real-world digital twin scenes, 8 high-level tasks, and 2.56M meters of trajectories, and is built on an aligned 3D Gaussian Splatting (3DGS)--Mesh hybrid representation that combines photorealistic rendering with collision-capable geometry for scalable generation and collision-aware evaluation. 
We introduce process-oriented and collision-aware metrics to assess process fidelity and flight safety. Experiments on representative state-of-the-art VLA models reveal significant gaps in high-level semantic completion and safe execution, highlighting HUGE-Bench as a diagnostic testbed for high-level UAV autonomy.
\end{abstract}

\begin{figure}[tb]
  \centering
  \includegraphics[width=\linewidth]{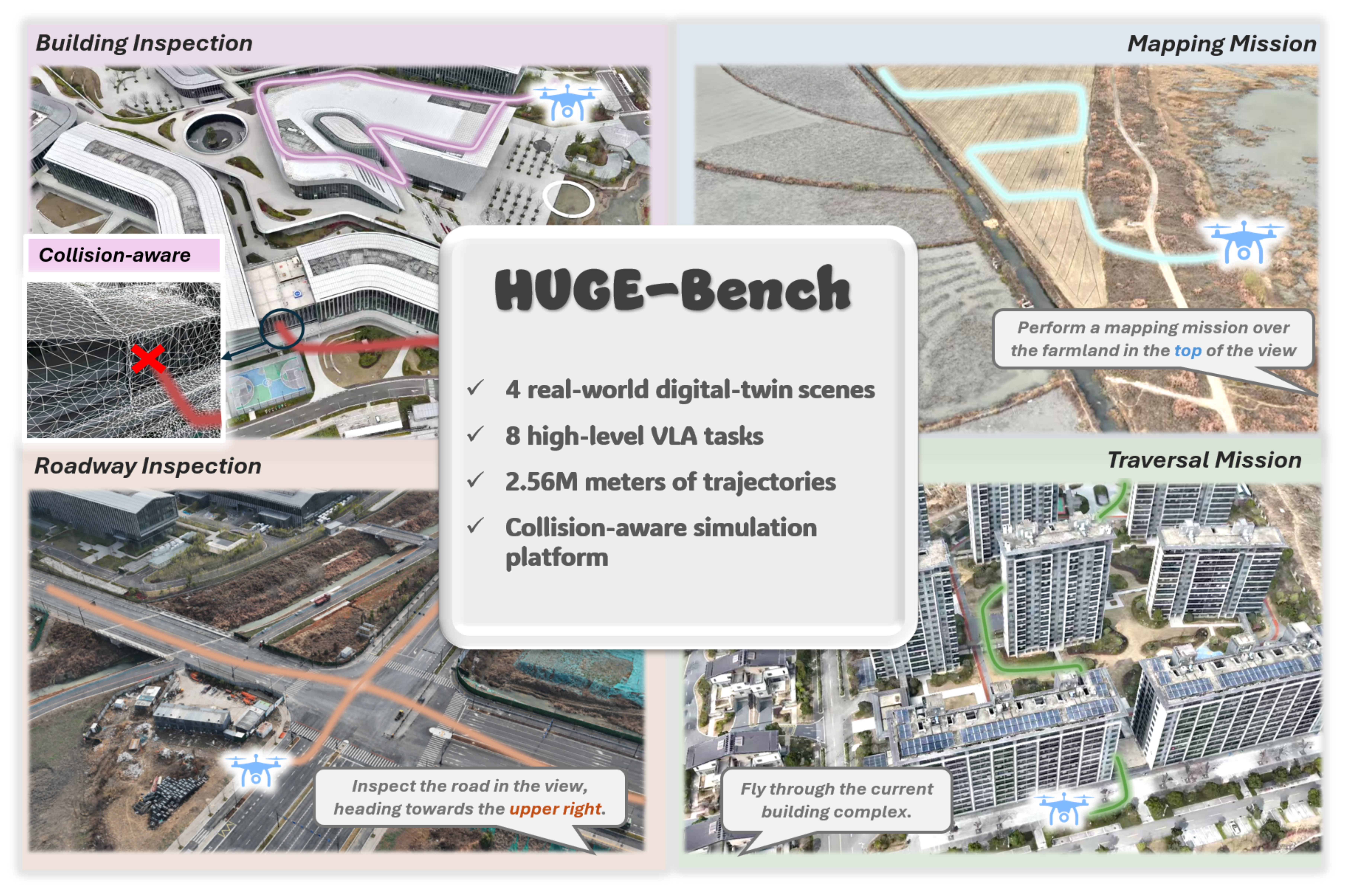}
  \caption{\textbf{Overview of HUGE-Bench.} Our benchmark comprises four 3DGS–Mesh digital-twin environments reconstructed from real-world scenes, eight high-level UAV VLA tasks, and 2.56 million meters of trajectory data. In addition, we develop a collision-aware simulation platform that enables collision-free trajectory collection and facilitates safety evaluation.
  }
  \label{fig1}
\end{figure}

\begin{table*}[t]
\centering
\fontsize{7.8pt}{8.8pt}\selectfont
\setlength{\tabcolsep}{4.0pt}
\renewcommand{\arraystretch}{1.2}
\caption{\textbf{Comparison of UAV navigation and action benchmarks. }
“Source” indicates whether the environments are synthetic or reconstructed from real-world data. 
“3D Rep.” denotes the 3D scene representation. 
“Subtask” indicates whether the benchmark provides explicit
multi-stage task decomposition rather than single-step navigation.
“Len. (m/w)” measures trajectory length normalized by instruction length.
“Collision” indicates explicit collision-aware evaluation.}
\label{tab:uav_benchmark_comparison}
\begin{tabular}{lcccccc}
\toprule
\textbf{Benchmark} & \textbf{Source} & \textbf{3D Rep.} & \textbf{Subtask} & \textbf{Len. (m/w)} & \textbf{Collision} & \textbf{Type} \\
\midrule
AerialVLN~\cite{liu2023aerialvln}   & Syn.        & Mesh        & $\times$ & \,\,\,7.97  & $\times$ & VLN \\
CityNav~\cite{lee2024citynav}     & Real        & P. Cloud    & $\times$ & 20.92 & $\times$ & VLN \\
TravelUAV~\cite{wang2024realisticuavvln}   & Syn.        & Mesh        & $\times$ & \,\,\,2.45  & $\checkmark$ & VLN \\
Openfly~\cite{gao2025openfly}     & Syn., Real  & Mesh, 3DGS  & $\times$ & \,\,\,1.68  & $\times$ & VLN \\
UAV-Flow~\cite{wang2025uavflow}    & Syn., Real  & Mesh        & $\times$ & \,\,\,\,$<4$  & $\times$ & VLA \\
\midrule
\textbf{HUGE-Bench} & \textbf{Real} & \textbf{3DGS-Mesh} & \textbf{$\checkmark$} & \textbf{32.07} & \textbf{$\checkmark$} & \textbf{VLA} \\
\bottomrule
\end{tabular}
\end{table*}

\newcommand{\cmark}{\ding{51}}
\newcommand{\xmark}{\ding{55}}

\begin{figure}[tb]
  \centering
  \includegraphics[height=6.5cm]{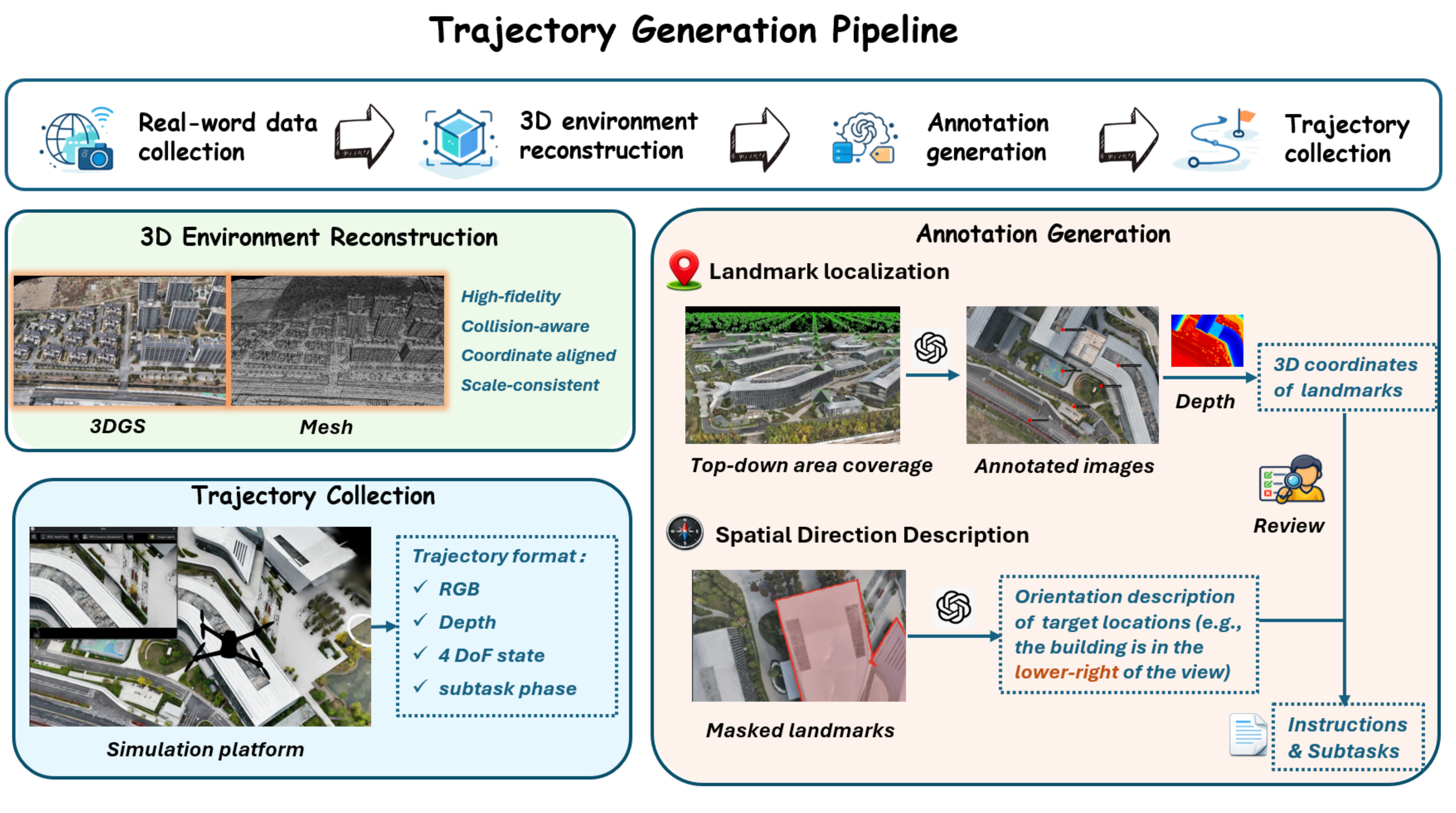}
  \caption{
\textbf{Trajectory generation pipeline of HUGE-Bench.} 
Starting from real-world data collection, we reconstruct 3D environments using 3DGS and mesh representations. 
Landmarks are localized and annotated to generate spatial descriptions and task instructions. Trajectories are collected in the simulation platform with multimodal observations, including RGB, depth, 4-DoF state, and subtask phases.
  }
  \label{fig2}
\end{figure}

\section{Introduction}
Unmanned aerial vehicles (UAVs) are increasingly deployed for inspection, search-and-rescue, infrastructure monitoring, and logistics, yet operating them in cluttered 3D environments remains labor-intensive and error-prone.
Operators must translate high-level intent into dense waypoints and continuous low-level control while maintaining situational awareness and safety.
Recent advances in vision-language-navigation (VLN) suggest a promising alternative: language-guided flight, where users express intent in natural language and an agent grounds it into perception and actions \cite{anderson2018r2r,krantz2020vlnce,liu2023aerialvln,gao2025openfly}.
In the UAV domain, aerial VLN benchmarks have further catalyzed progress by standardizing tasks, datasets, and evaluation protocols \cite{liu2023aerialvln,gao2025openfly}.
However, existing benchmarks still only partially reflect how UAVs are commanded and evaluated in real operations.

A central mismatch lies in instruction style and task abstraction.
Most VLN settings emphasize long, step-wise route descriptions, and evaluation is largely goal-centric, e.g., Success Rate (SR) and Success weighted by Path Length (SPL) \cite{anderson2018r2r}, sometimes complemented by normalized Dynamic Time Warping (nDTW) and Success weighted by nDTW (SDTW) for trajectory fidelity \cite{ilharco2019ndtw}.
In contrast, UAV operators typically issue brief, high-level commands (e.g., ``inspect the building on the left''), leaving the system to infer targets, decompose subtasks, and execute a multi-stage procedure safely.
Such commands stress semantic grounding from aerial viewpoints, 3D spatial reasoning over direction/altitude/distance, and memory across stages.
Yet most existing UAV benchmarks are still formulated as route-following VLN, which makes them less diagnostic for short, underspecified commands and process-oriented multi-stage behaviors (Table~\ref{tab:uav_benchmark_comparison}).

This gap is further amplified by environment representation and safety modeling.
Benchmark validity depends on both photorealistic perception and physically executable collision checking.
Mesh-based simulators support physics and collision queries (e.g., high-fidelity UAV simulation in AirSim) \cite{shah2017airsim}, but may introduce a perception gap for vision--language grounding.
Conversely, neural rendering such as 3D Gaussian Splatting (3DGS) provides real-time, photorealistic novel-view rendering \cite{kerbl2023gaussiansplatting}, but native 3DGS is appearance-first and does not directly yield collision-ready geometry.
Recent efforts toward physically executable 3DGS highlight the need for hybrid designs that pair 3DGS rendering with mesh-based collision bodies \cite{miao2025sage3d}, which is particularly crucial for safety-critical UAV evaluation. Moreover, current evaluation protocols under-measure process correctness.
Goal-only metrics can miss important failure modes in high-level missions: a policy may reach an endpoint while skipping required procedures, drifting from the intended process, or colliding en route.
Many UAV missions are inherently process-oriented (inspection coverage, orbiting with clearance, mapping completeness), and thus require metrics that assess procedural completion beyond endpoint success.

To address these limitations, we introduce \textbf{HUGE-Bench}, a benchmark for High-Level UAV Vision Language Action (HL-VLA) control.
HUGE-Bench contains 4 real-world 3DGS-Mesh scenes, 8 high-level tasks, and 2.56M meters of trajectories, together with collision-aware simulation.
Unlike route-following VLN, HUGE-Bench targets the operational regime where a short command implies structured multi-stage behaviors.
Our task suite spans landing, road/building inspection, area mapping, orbiting (altitude/radius control), multi-turn spiral-down, and obstacle-aware region traversal (Fig.~\ref{fig1}). \textbf{HUGE-Bench} is enabled by a real-to-sim pipeline that couples photorealistic rendering with physics-grounded collision evaluation (Fig.~\ref{fig2}).
We reconstruct an aligned 3DGS--Mesh digital twin: 3DGS supports realistic perception, while the mesh provides collision queries and depth.
This design supports scalable trajectory generation while preserving an explicit safety channel (collision outcomes), enabling evaluation of both semantic grounding under short commands and safe execution.

We propose evaluation tailored to high-level UAV behaviors.
We report Average Trajectory Coverage Rate (Avg. TCR), normalized Dynamic Time Warping (nDTW), and Normalized Task Progress (NTP) to assess process completion, trajectory alignment, and task progress. For safety-critical traversal, we additionally measure Collision Rate (CR) and Collision-aware Success weighted by Path Length (CSPL) to evaluate collision frequency, goal reaching, and path efficiency.
Using this protocol, we benchmark representative VLA and VLM models (OpenVLA, FastVLM, $\pi_0$, and $\pi_{0.5}$) \cite{kim2024openvla,vasu2025fastvlm,pi2025pi0,pi2025pi05}.
Results reveal substantial gaps in process completion and safe execution under short instructions, positioning HUGE-Bench as a diagnostic and safety-relevant testbed for high-level UAV VLA.

In summary, our contributions are three-fold:
\begin{itemize}
    \item \textbf{HL-VLA task formulation.} We introduce a new benchmark setting for HL-VLA tasks, where UAVs must interpret short, potentially ambiguous commands and execute multi-stage semantic behaviors.

    \item \textbf{Real-to-sim benchmark.} We build HUGE-Bench, a UAV benchmark constructed from real-world scenes and an aligned 3DGS--Mesh digital twin, enabling large-scale trajectory generation and realistic safety-aware evaluation.

    \item \textbf{Process-oriented and safety evaluation.} We propose process-oriented and collision-aware metrics that evaluate UAV execution along three dimensions: process fidelity and safety.
\end{itemize}

\section{Related Work}

\subsection{Aerial Vision--Language Navigation and VLA}
Vision--Language Navigation (VLN) studies instruction-conditioned navigation, typically evaluated with goal-centric success/efficiency metrics (e.g., SR/SPL) and trajectory-faithfulness scores (e.g., nDTW/SDTW) \cite{anderson2018r2r,krantz2020vlnce,ilharco2019ndtw}.
Large-scale VLN datasets and variants (e.g., R4R, RxR, CVDN, REVERIE) further expose long-horizon and grounded language challenges \cite{jain2019r4r,ku2020rxr,thomason2020cvdn,qi2020reverie}.
Methodologically, VLN has progressed from imitation/RL baselines to pretraining and transformer-based policies with longer memory and stronger grounding \cite{hao2020prevalent,chen2021hamt,chen2022duet,gu2022vlnsurvey}.

Aerial VLN inherits this protocol but must handle continuous 3D flight dynamics, larger scale, and stronger safety constraints.
AerialVLN proposes Look-ahead Guidance (LAG) to mitigate supervision mismatch between shortest paths and language-described routes \cite{liu2023aerialvln}.
OpenFly scales aerial VLN via automated trajectory--instruction generation across multiple engines and introduces a keyframe-aware agent with history and token merging \cite{gao2025openfly}.
Recent VLM/VLA-based navigation explores video-conditioned action prediction and cross-embodiment generalization \cite{zhang2024navid,zhang2025uninavid,cheng2024navila}.
Despite these advances, the dominant formulation remains route-following with detailed instructions.
In contrast, we study high-level UAV Vision Language Action tasks under short, underspecified commands, where success requires process-oriented multi-stage execution and safety-aware behavior.

\subsection{UAV VLN/VLA Benchmarks}
UAV benchmarks vary widely in data sources, scene representations, annotation granularity, and safety protocols.
As summarized in Table~\ref{tab:uav_benchmark_comparison}, AerialVLN \cite{liu2023aerialvln}, CityNav \cite{lee2024citynav}, and OpenFly \cite{gao2025openfly} are primarily VLN-style benchmarks without intermediate subtask annotations and with limited collision-centric evaluation, while TravelUAV\cite{wang2024realisticuavvln} emphasizes more realistic UAV simulation and assistant-guided object search.
UAV-Flow shifts to language-conditioned fine-grained control with real-world episodes, complementing VLN with a VLA-style control regime \cite{wang2025uavflow}.
Overall, existing resources underrepresent the regime of short language driving long, multi-stage flights and rarely standardize collision-aware evaluation.
HUGE-Bench is designed to occupy this missing region with (i) high-level task definitions, (ii) scalable trajectory generation, and (iii) explicit collision-aware evaluation enabled by our digital twin.

\subsection{Environment Representations and Evaluation}
Safety-relevant benchmarking depends on both photorealistic perception and physically executable collision checking.
UAV simulation platforms such as AirSim support physics-based UAV dynamics and collision queries \cite{shah2017airsim}, while embodied AI simulators and datasets (e.g., Habitat, Matterport3D, HM3D) provide efficient rendering and large-scale reconstructed geometry \cite{savva2019habitat,chang2017matterport3d,ramakrishnan2021hm3d}.
Neural rendering, especially 3D Gaussian Splatting (3DGS), enables high-quality real-time novel-view synthesis \cite{kerbl2023gaussiansplatting}, but is not collision-ready by default; recent work therefore advocates hybrid designs that pair 3DGS rendering with mesh-based collision bodies for physically executable evaluation \cite{miao2025sage3d}.
On the metrics side, classic VLN emphasizes SR/SPL and trajectory similarity (nDTW/SDTW), and complementary measures such as instruction-fidelity/coverage have also been proposed \cite{anderson2018r2r,ilharco2019ndtw,jain2019r4r}.
Motivated by the need to evaluate process correctness and safety in high-level UAV behaviors, HUGE-Bench adopts an aligned 3DGS--Mesh digital twin and reports process-oriented and collision-aware metrics tailored to multi-stage trajectories rather than goal-only navigation.

\begin{figure}[tb]
  \centering
  \includegraphics[width=\linewidth]{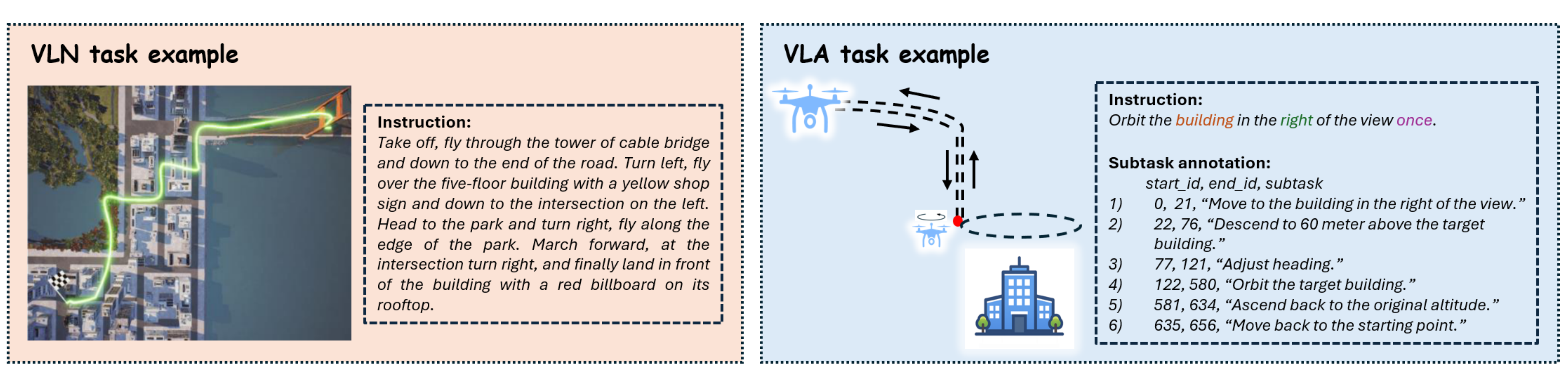}
    \caption{\textbf{Difference between previous VLN tasks and HL-VLA tasks.} 
    Left: a VLN task example~\cite{liu2023aerialvln} where the UAV follows a long-horizon navigation instruction through multiple landmarks. 
    Right: a HL-VLA task example with a high-level instruction decomposed into temporally aligned subtasks for structured trajectory execution.}
  \label{fig3}
\end{figure}

\section{HUGE-Bench}
\label{sec:huge_bench}

We introduce HUGE-Bench, a benchmark for high-level Vision--Language--Action control of UAVs. HUGE-Bench includes (i) a task suite with representative high-level UAV behaviors, (ii) a scalable trajectory and annotation generation pipeline grounded in real-world capture, (iii) collision-aware simulation and data generation platform based on an aligned 3DGS--Mesh digital twin, and (iv) process-oriented evaluation metrics that jointly measure process fidelity and safety. An overview of the pipeline is shown in Fig.~\ref{fig2}.

\subsection{High-level VLA Tasks for UAVs}
\label{subsec:tasks}
We consider a UAV VLA policy that, given the current observation (e.g., RGB or RGBD) and a natural-language instruction, predicts the next-step action target, such as velocity commands or the next state along a trajectory:
\begin{equation}
\mathbf{a}_t = \pi_\theta\!\left(\mathbf{o}_t, \mathbf{x}, \mathbf{s}_t\right),
\label{eq:vla_policy}
\end{equation}
where $\mathbf{o}_t$ denotes the UAV observation at time $t$, $\mathbf{x}$ is the instruction, $\mathbf{s}_t$ is the current state, and $\mathbf{a}_t$ is the action target (e.g., velocity $\mathbf{v}_t$ or the next state $\mathbf{s}_{t+1}$) predicted by the policy $\pi_\theta$ with parameters $\theta$.

\textbf{High-level instructions} in our setting are short and potentially ambiguous, reflecting realistic human interaction patterns (e.g., ``inspect the building on the left''). Unlike step-by-step low-level commands, a high-level instruction typically implies a sequence of latent subtasks. For example, ``inspect the building on the left'' requires the UAV to: (1) identify the referred building from an aerial viewpoint, (2) approach it, (3) descend to an inspection altitude, (4) orbit while maintaining safe clearance, and (5) return to the initial altitude and position. Therefore, HL-VLA for UAVs stresses: (1) Language grounding and subtask decomposition under short and underspecified instructions, (2) Semantic visual understanding from aerial perspectives, and (3) 3D spatial reasoning (direction, altitude, distance, orientation). A comparison between conventional UAV VLN and our proposed HL-VLA task is illustrated in Fig.~\ref{fig3}

\begin{figure}[tb]
  \centering
  \includegraphics[width=\linewidth]{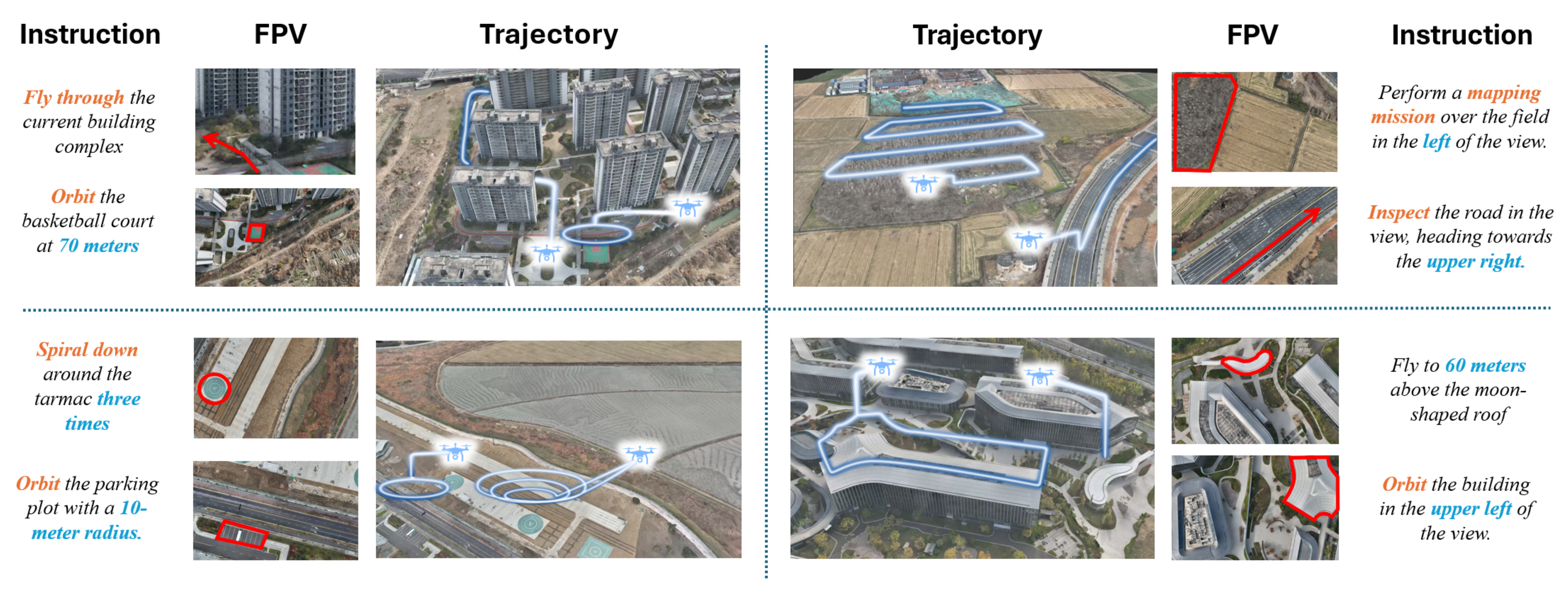}
\caption{\textbf{Examples of HL-VLA tasks in HUGE-Bench.} 
Each example shows a natural-language instruction, the corresponding first-person view (FPV) observations highlighting target regions, and the resulting multi-stage flight trajectory. }
  \label{fig4}
\end{figure}

Based on these requirements, we define eight representative tasks (see Fig.~\ref{fig4}):
\begin{enumerate}
    \item Target Landing: orient toward a target, reach its overhead position, descend to a specified altitude, and hover \,(referred to as \textbf{Landing}).
    \item Road Inspection: reach the road, descend to inspection altitude, align with the required direction, perform inspection, then ascend back to the start altitude and stop \,(referred to as \textbf{Inspection-R}).
    \item Adaptive Building Inspection: approach the building boundary, descend to inspection altitude, select an orbit direction, circumnavigate along the boundary with adaptive clearance, then return to the start pose \,(referred to as \textbf{Inspection-B}).
    \item Area Mapping: move to the boundary of a target region, descend to a specified altitude, perform coverage-style mapping, and return to the start altitude \,(referred to as \textbf{Mapping}).
    \item Orbiting at Different Heights: reach the target overhead location at a specified altitude, enter a circular track, loiter, then exit and return \,(referred to as \textbf{Orbit-H}).
    \item Orbiting with Different Radius: similar to 5. but with an instruction-conditioned orbit radius \,(referred to as \textbf{Orbit-R}).
    \item Multi-turn Spiral Down: execute a spiral loiter pattern (multi-turn) conditioned on the instruction \,(referred to as \textbf{Spiral Down}).
    \item Region Traversal: traverse a designated area while detecting obstacles and executing avoidance behaviors \,(referred to as \textbf{Traversal}).
\end{enumerate}
Collectively, these tasks probe target recognition, directional grounding, depth and altitude awareness, and multi-stage trajectory completion in a unified HL-VLA setting.

\subsection{Trajectory Data Generation}
\label{subsec:data_generation}

HUGE-Bench trajectories are generated through a real-to-sim pipeline that couples photorealistic perception with physics-grounded collision awareness. The pipeline consists of: (1) real-world data capture, (2) aligned 3DGS--mesh digital twin reconstruction, (3) annotation generation, and (4) simulator-based trajectory collection.

\subsubsection{Real-world Data Capture.}
\label{subsubsec:capture}

We collect data using a DJI M400 UAV equipped with a Zenmuse L2 payload across four representative outdoor scenes: office buildings, dense urban blocks, swamp and farmland, and construction roads. The total covered area is approximately $6.45~\mathrm{km}^2$. Flights are performed at $\sim 85~\mathrm{m}$ above ground, recording GPS from GNSS, UAV pose, camera pose, and high-resolution RGB images ($5280\times 3956$). To improve reconstruction quality at near-ground altitudes, we additionally capture supplementary low-altitude views.

\subsubsection{3D Environment Reconstruction.}
\label{subsubsec:recon}

For each scene, we reconstruct both a 3DGS model and a triangle mesh. This joint 3DGS--Mesh digital twin representation is central to HUGE-Bench:
3DGS provides high-fidelity renderings for realistic perception inputs.
The mesh provides collision-enabled geometry and supports physics-based queries, as well as depth via ray casting.

\subsubsection{Landmark Localization.}
We place a grid of top-down cameras at $\sim 60~\mathrm{m}$ altitude and render the 3DGS scene to obtain global 2D map views, while rendering the mesh to obtain aligned depth. We then use an LLM to localize landmarks on the rendered maps and output normalized image coordinates. After manual review to remove ambiguous or incorrect labels, we back-project pixel coordinates using camera intrinsics and depth to obtain pixel-level 3D landmark positions. For semantically extended regions (e.g., buildings, roads, fields), we additionally provide human-annotated masks.

\subsubsection{Spatial Direction Descriptions.}
Many HL tasks cannot be reliably specified with only generic nouns like ``building'' or ``road,'' especially when multiple visually similar candidates appear in view or when a route direction must be specified (e.g., road inspection ``along the right side''). To address this, before trajectory collection we select the intended target, sample random initial views in its vicinity, and use the target mask to prompt an LLM to produce human-like spatial referring expressions (e.g., ``upper-left,'' ``center,'' ``along the right side of the frame''). This yields instructions combining directional phrases with generic categories (e.g., ``the building in the upper-left''), enabling more precise language grounding under high-level commands.

\subsubsection{Simulator-based Trajectory Collection}
\label{subsubsec:sim}

Trajectory collection is conducted in Isaac Sim~\cite{isaacsim}, which offers GPU-accelerated rendering and PhysX-based rigid-body simulation. First, we import the aligned 3DGS--Mesh digital twin into the simulator. Then we generate a sequence of 3D waypoints using task-specific rules. During execution, the simulator records synchronized multimodal streams including RGB, depth, pose, flight states, and collision signals, enabling scalable data generation and standardized evaluation. For data sampling, we record frames at approximately $1~\mathrm{m}$ spatial intervals along the trajectory; during turns, we additionally sample at every $5^\circ$ heading change to better capture viewpoint transitions. For the obstacle-avoidance traversal task, we generate paths using an RRT-based planner~\cite{karaman2011anytime}. Because dense triangle meshes can significantly increase planning and collision-query cost, we perform mesh cleaning floater removal and mesh decimation to reduce complexity. Additional implementation details are provided in the Appendix.

\begin{figure}[tb]
  \centering
  \includegraphics[width=\linewidth]{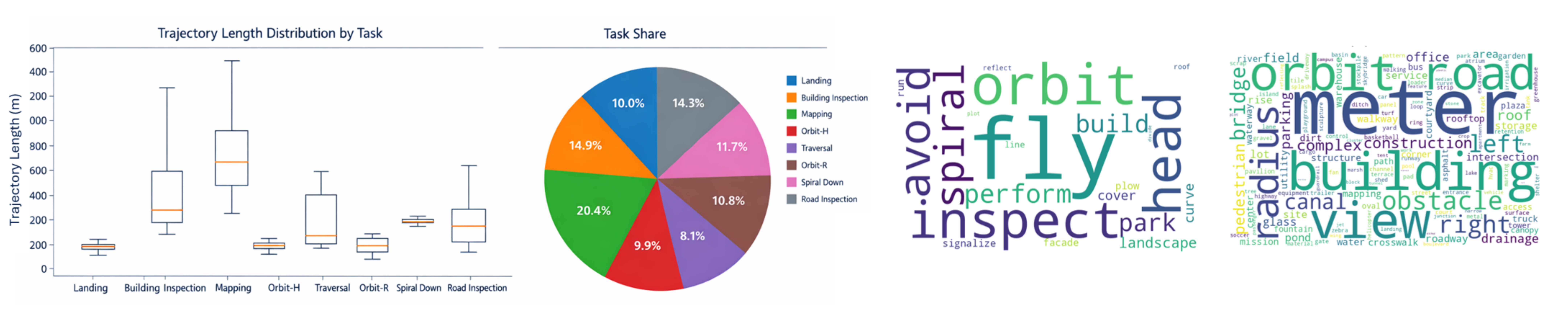}
\caption{\textbf{Dataset statistics of HUGE-Bench.} 
We analyze trajectory length distributions across tasks (left), the proportion of different task types (middle), and the vocabulary distribution in instructions through verb and landmark word clouds (right).}
  \label{fig5}
\end{figure}

\subsection{Data Analysis}
\label{subsec:data_analysis}

Fig.~\ref{fig5} summarizes the dataset statistics. Trajectory lengths vary substantially across tasks: Mapping exhibits the longest and most diverse trajectories, while Landing, Orbit-H, Orbit-R, and Spiral Down are shorter and more concentrated. The task distribution is relatively balanced, with Mapping as the largest subset (20.4\%) and Traversal the smallest (8.1\%). The two word clouds summarize the dominant semantics in HL-VLA instructions. The left cloud is dominated by scene entities and landmarks (e.g., building, road, tree, obstacle). The right cloud highlights action-oriented verbs (e.g., fly, orbit, avoid, inspect, spiral), suggesting that the tasks emphasize high-level motion primitives and inspection behaviors beyond simply reaching a goal.

\subsection{Dataset Splits}
We partition our dataset into three splits: train, test-seen, and test-unseen. This design aims to evaluate a model's ability to generalize across both outdoor environments and natural-language instructions. 

In our benchmark, we define test-seen as trajectories whose target landmark appears in the training set. Although the landmark itself is seen during training, the initial observation in the first frame is randomized, including the agent's altitude, horizontal distance to the target, and the relative direction with respect to the landmark. This split evaluates whether the model can robustly reach a familiar target under novel initial viewpoints and spatial configurations. 

To further assess generalization to new scenarios and language, we construct a test-unseen split from two complementary sources:
(1) Landmark-unseen: we hold out a subset of landmarks that never appear in training.
(2) Instruction-unseen: we apply instruction-level generalization by randomly diversifying the language. Specifically, we use an LLM to generate paraphrases that replace the verb or the textual description of the target landmark while preserving the intended task semantics. In total, our dataset contains: 5{,}330 trajectories in train, 593 trajectories in test-seen, and 294 trajectories in test-unseen. This split measures the model's ability to handle novel landmarks and linguistic variations beyond the training distribution.

\subsection{Evaluation Metrics}
\label{subsec:metrics}

Classic goal-oriented VLN benchmarks mainly emphasize terminal success. In contrast, our HL-VLA suite requires evaluating process completion, trajectory alignment and safety-aware execution. For process-oriented tasks such as inspection, mapping, and orbiting, we use \textbf{Trajectory Coverage Rate (TCR)} to measure how well a predicted trajectory covers the intended ground-truth process. Given a ground-truth trajectory $T^{gt}=\{p_i\}_{i=1}^{N}$ and a predicted trajectory $T^{pred}$ treated as a polyline, we compute:
\begin{equation}
d_i = \min_{q \in T^{pred}} \|p_i - q\|_2 .
\end{equation}
TCR is then defined as:
\begin{equation}
\mathrm{TCR} = \frac{1}{N}\sum_{i=1}^{N}\mathbb{I}(d_i < \delta),
\end{equation}
where $\delta$ is a distance tolerance. We report $\mathrm{TCR}@K$ under multiple thresholds, where $K \in \{1\,\mathrm{m}, 2\,\mathrm{m}, 5\,\mathrm{m}\}$. We further report \textbf{Avg. TCR}, which averages TCR scores over the evaluated tasks to summarize overall process completion.

Since TCR measures spatial coverage but is less sensitive to temporal ordering, we also report \textbf{normalized Dynamic Time Warping (nDTW)}~\cite{ilharco2019ndtw}. nDTW penalizes reversed or re-ordered trajectories, and therefore complements TCR by measuring direction-aware trajectory alignment.

We further report \textbf{Normalized Task Progress (NTP)} to measure how much of the intended high-level task is completed. Unlike terminal success, which only evaluates whether the final goal is reached, NTP provides a normalized progress score in $[0,1]$ based on the completion of annotated task stages. This allows us to evaluate partial task completion even when the full trajectory is not successfully completed.

For safety-critical traversal, we compute \textbf{Collision Rate (CR)} as the fraction of episodes in which at least one collision occurs, following common practice in collision-aware simulation benchmarks~\cite{puig2024habitat3}. To jointly reflect goal reaching, collision-free execution, and path efficiency, we further report \textbf{Collision-aware SPL (CSPL)} based on Success weighted by Path Length (SPL)~\cite{anderson2018spl}:
\begin{equation}
\mathrm{CSPL} = S \cdot (1-\mathbb{I}(C>0)) \cdot \frac{L}{\max(P, L)},
\end{equation}
where $S$ indicates terminal success, $L$ is the reference path length, and $P$ is the predicted path length. Here, $\mathbb{I}(C>0)$ is an episode-level collision indicator that equals $1$ if any collision occurs and $0$ otherwise.

Overall, these metrics capture trajectory alignment and task progress (TCR, nDTW and NTP), and safety-aware efficiency (CR and CSPL), enabling a faithful evaluation of HL-VLA behaviors.

\begin{table}[tb]
\centering
\small
\setlength{\tabcolsep}{6pt}
\renewcommand{\arraystretch}{0.95}
\caption{
\textbf{Overall comparison on the test set.}
We report both process-oriented metrics and safety metrics across different baseline models. 
\textcolor{lightgraymetric}{Gray} values indicate limited comparative significance due to low performance, e.g., spinning in place.
}
\label{tab:overall_process_safety}
\resizebox{\columnwidth}{!}{%
\begin{tabular}{lccccc}
\toprule
\multirow[c]{2}{*}{\textbf{Methods}}
& \multicolumn{3}{c}{\textbf{Process-oriented Metrics}}
& \multicolumn{2}{c}{\textbf{Safety Metrics}} \\
\cmidrule(lr){2-4}\cmidrule(lr){5-6}
& Avg. TCR $\uparrow$
& nDTW $\uparrow$
& NTP $\uparrow$
& CR $\downarrow$
& CSPL $\uparrow$ \\
\midrule
OpenVLA
& 0.112
& 0.011
& 0.122
& \textcolor{lightgraymetric}{0.003}
& \textcolor{lightgraymetric}{0.032} \\

MemoryVLA
& 0.234
& 0.077
& 0.549
& 0.072
& 0.226 \\

FastVLM
& 0.285
& 0.136
& 0.633
& 0.060
& 0.391 \\

$\pi_0$
& 0.558
& 0.443
& \textbf{0.653}
& \textbf{0.011}
& 0.802 \\

Depth-aware $\pi_{0.5}$
& 0.577
& 0.459
& 0.617
& 0.013
& 0.792 \\

$\pi_{0.5}$
& \textbf{0.581}
& \textbf{0.467}
& 0.618
& 0.018
& \textbf{0.805} \\
\bottomrule
\end{tabular}%
}
\end{table}

\begin{figure}[tb]
  \centering
  \includegraphics[width=\linewidth]{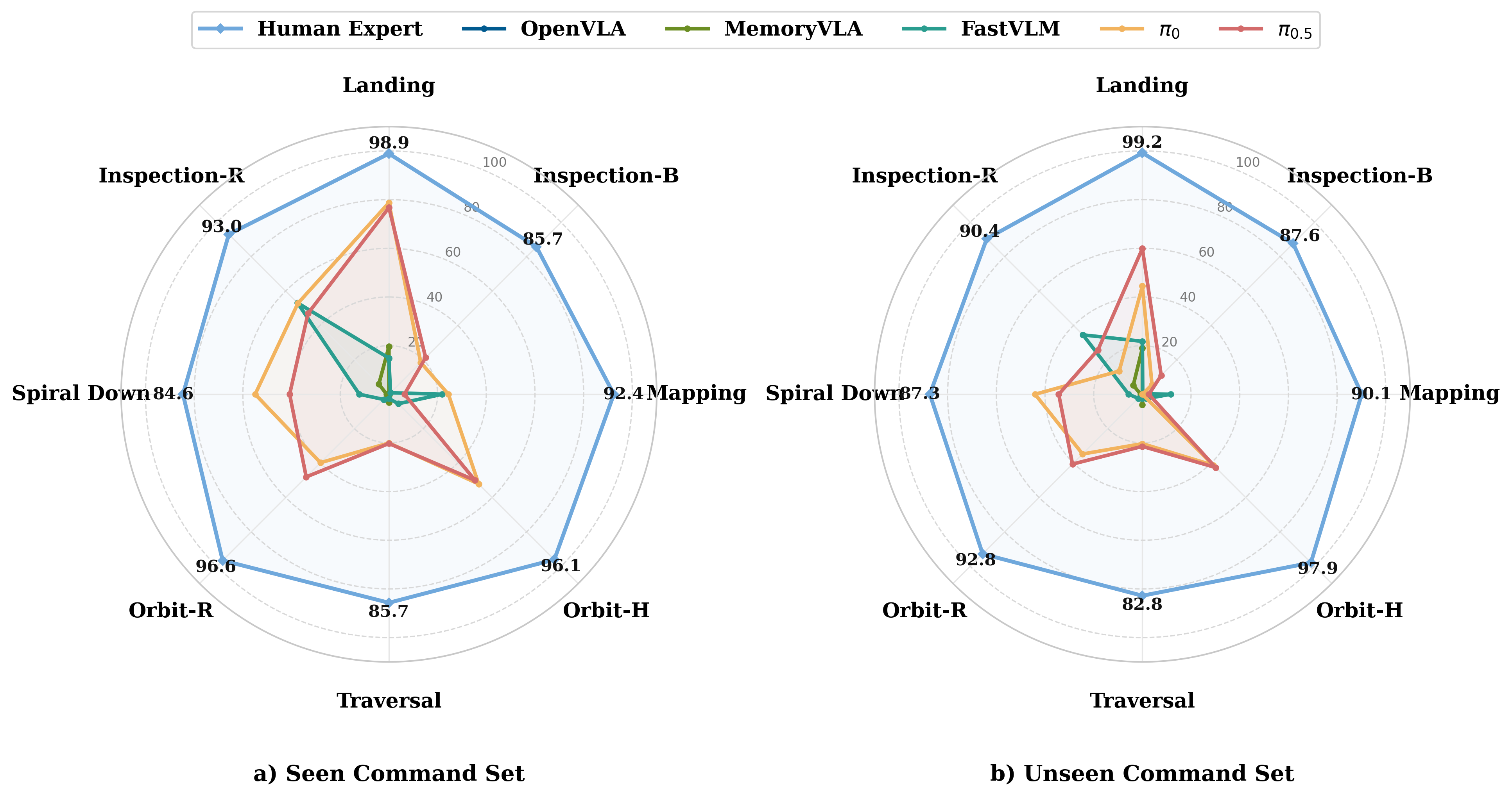}
    \caption{Per-task nDTW comparison on seen and unseen sets.
    }
  \label{fig_per}
\end{figure}

\section{Experiments}
\subsection{Baselines}
The HL-VLA task requires strong vision--language understanding, reliable spatial perception, and memory over multi-stage execution. We therefore evaluate representative state-of-the-art VLA models on HUGE-Bench, including OpenVLA~\cite{kim2024openvla}, $\pi_{0}$~\cite{pi2025pi0}, $\pi_{0.5}$~\cite{pi2025pi05}, and FastVLM~\cite{vasu2025fastvlm}. These baselines cover a trade-off between general-purpose capability and efficiency. For FastVLM, we use FastVLM to efficiently encode visual and language inputs into hidden states, and append $\pi_{0}$'s action expert, initialized from $\pi_{0}$, to predict actions.

We additionally evaluate MemoryVLA, an OpenVLA-based baseline with implicit memory, to examine whether memory improves long-horizon HL-VLA execution. To study multimodal spatial perception, we also include a depth-aware $\pi_{0.5}$ baseline, which encodes depth maps as additional visual observations.

Following prior work~\cite{wang2025uavflow}, we provide the model with visual observations from two time steps: the first frame of the trajectory and the current frame. This two-frame input is intended to help the model infer the agent's relative position and progress in 3D space. To ensure a fair comparison, all baselines are evaluated under the same input format and protocol on HUGE-Bench. Additional details on the models, training procedures, and implementation settings are provided in the Appendix.

\subsection{Results}

Table~\ref{tab:overall_process_safety} summarizes the overall performance on the test set. 
Overall, the $\pi$-based policies achieve the strongest performance among the evaluated baselines. 
$\pi_{0.5}$ obtains the best Avg. TCR, nDTW, and CSPL, suggesting that richer visual pretraining can improve overall trajectory quality and process alignment. 
$\pi_0$ achieves the best NTP and the lowest CR, indicating stronger stage-wise progress and safer execution in our evaluation. 
FastVLM remains competitive on process-oriented metrics, while OpenVLA performs substantially worse.
MemoryVLA improves over OpenVLA, confirming the benefit of memory for long-horizon HL-VLA tasks. 
This supports the need for future memory-augmented VLA models that can better track task progress across multi-stage aerial behaviors.

To evaluate safety behavior, we further study collision-aware metrics enabled by our 3DGS--Mesh digital twin. 
As shown in Table~\ref{tab:overall_process_safety}, $\pi_0$ achieves the lowest Collision Rate, while $\pi_{0.5}$ obtains the best collision-aware path quality measured by CSPL. 
The depth-aware $\pi_{0.5}$ baseline reduces CR from 0.018 to 0.013 compared with $\pi_{0.5}$, but does not consistently improve trajectory alignment or overall task progress. 
These results suggest that richer visual pretraining and explicit depth can improve different aspects of behavior, but neither alone guarantees safe and reliable 3D execution.

Fig.~\ref{fig_per} provides a per-task nDTW comparison on seen and unseen splits. 
Across most tasks, the $\pi$-based policies maintain stronger trajectory alignment than OpenVLA and FastVLM, especially on landing and orbiting tasks where landmark recognition and localization are central. 
The seen--unseen gap is task-dependent: models generally transfer better on structured landmark-centric behaviors, while inspection, mapping, spiral-down, and traversal remain more challenging because they require sustained geometric tracking, long-horizon progress estimation, and safe 3D execution. 
These results show that HUGE-Bench covers a broad difficulty spectrum rather than only measuring endpoint reaching. To contextualize model performance, we further include expert-pilot evaluation as a human upper bound, and implementation details are provided in the supplementary material.

\begin{figure}[tb]
\centering
\includegraphics[width=\linewidth]{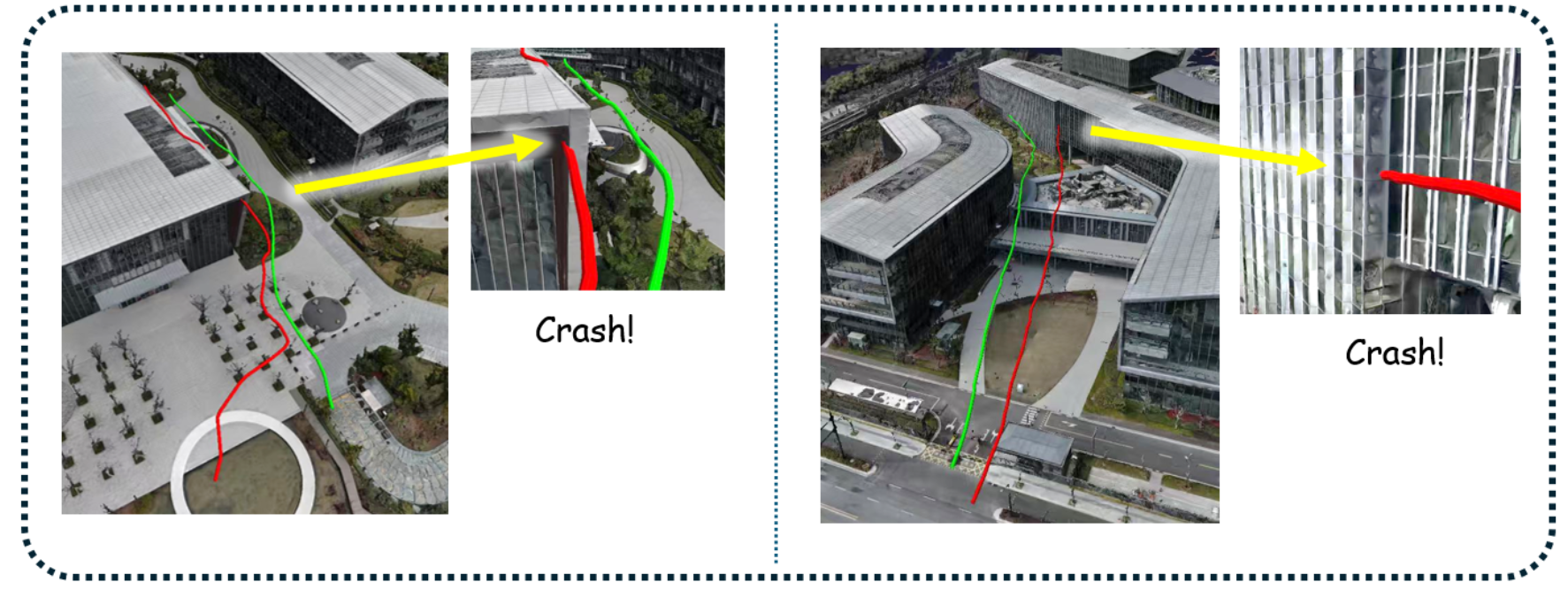}
\caption{\textbf{A visualization of model predictions for Traversal tasks.} The green trajectories indicate collision-free results, and the red trajectories denote trajectories that result in collisions.}
\label{fig_t}
\end{figure}

\begin{figure}[tb]
  \centering
  \includegraphics[width=\linewidth]{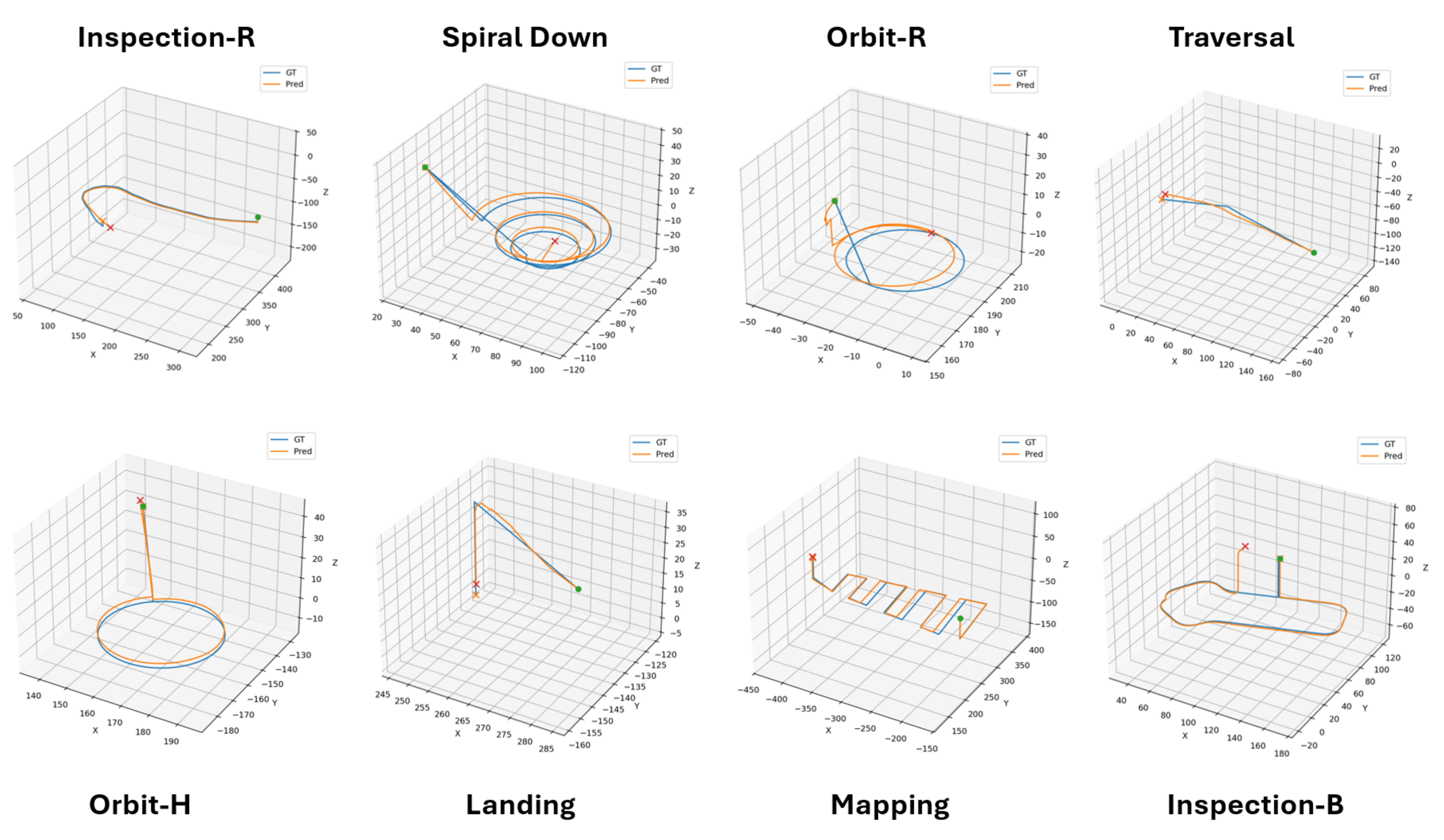}
\caption{\textbf{Visualization of successful results from $\pi_0$ across different tasks.} }
  \label{fig6}
\end{figure}

\section{Conclusion}
We introduce HUGE-Bench, a benchmark for evaluating high-level UAV vision--language--action capabilities in long-horizon missions. HUGE-Bench targets a practical operating regime in which a UAV must execute multi-stage tasks from brief, high-intent instructions while maintaining semantic correctness and safety. Unlike conventional UAV-VLN settings that primarily emphasize reaching a destination, our benchmark reflects a more realistic paradigm: short instruction $\rightarrow$ implicit subtask decomposition $\rightarrow$ stage-wise execution. To enable fine-grained diagnosis, we incorporate explicit stage annotations and process-oriented metrics to quantify execution progress and stage completion.

To support scalable and cost-effective evaluation, we build an end-to-end data and assessment pipeline that combines real UAV capture with an aligned 3DGS--Mesh digital twin. This hybrid representation leverages the high-fidelity rendering of 3DGS and the collision-aware geometry of meshes, enabling physically actionable simulation with collision-sensitive assessment as well as large-scale data generation. \textsc{HUGE-Bench} further releases richly annotated episodes, including RGB, depth, pose, destination semantics, and subtask stage labels. We also propose a unified evaluation protocol that jointly characterizes stage completion, trajectory efficiency, and collision safety. Experiments with representative state-of-the-art VLA models reveal substantial gaps in high-level semantic stage completion and safe execution, highlighting the value of HUGE-Bench for diagnosing limitations and advancing high-level UAV VLA systems.

\subsubsection{Limitations} 
The current benchmark primarily focuses on static environments. Incorporating diverse and realistic dynamics, e.g., moving obstacles, changing illumination and weather, or non-stationary objects, remains an important direction to better reflect real-world deployment conditions. Furthermore, while our digital-twin pipeline improves physical plausibility and enables collision-aware evaluation, bridging the gap from complex long-horizon high-level VLA behaviors in simulation to robust real-world execution is still challenging. Future work should investigate stronger domain generalization, online adaptation, and hardware-in-the-loop validation to facilitate reliable real-world deployment.

\clearpage
\appendix
\section*{Supplementary Material}
\label{sec:supplementary}

\section{Implementation Details}
\label{sec:implementation_details}
\subsubsection{3DGS--Mesh Reconstruction from Real-World Data.}
We collect large-scale outdoor aerial data using a DJI M400 UAV equipped with a Zenmuse L2 sensor. The captured data include high-resolution RGB images, calibrated camera poses, and GNSS/RTK measurements. We first process the raw data with an industrial aerial mapping pipeline to recover georeferenced camera poses and a metrically consistent mesh reconstruction. Based on the recovered poses, we further train a 3DGS model for photorealistic view synthesis. In terms of construction cost, each scene requires 8--10 hours of professional UAV flight and 12--15 hours of reconstruction and alignment on an RTX 5090 GPU workstation.

The mesh preserves scene geometry, physical boundaries, and is therefore used for depth rendering, collision checking, and motion planning. In contrast, 3DGS provides high-fidelity visual observations from arbitrary viewpoints. Since both representations are built from the same georeferenced acquisition and share the same coordinate system, they are naturally aligned in space. This joint 3DGS--Mesh representation forms the digital twin used in HUGE-Bench, enabling both realistic visual rendering and collision-aware physical simulation.

\subsubsection{Obstacle-Aware Path Planner.}
For obstacle-avoidance tasks, we generate trajectories using an RRT-based planner~\cite{karaman2011anytime} on the simplified mesh. To improve efficiency, we first remove floating artifacts and reduce mesh complexity, and then crop the planning mesh to a local region around the start--goal corridor. During planning, each candidate path segment is validated by collision and clearance checks against the mesh to ensure safe navigation. After a feasible path is found, we apply path simplification, smoothing, and fixed-interval resampling to obtain stable waypoints. For horizontal-view trajectories, we further decompose the motion into two stages: turn to face the target and fly to the target while avoiding obstacles. The turning stage is sampled at a fixed angular interval, while the flight stage uses smoothed headings along the planned path, resulting in visually consistent obstacle-avoidance trajectories.

\subsubsection{Rendering Details for Horizontal-View Data.}
For horizontal-view trajectories, directly rendering 3DGS often produces unrealistic empty backgrounds in regions not covered by scene geometry. To improve visual realism, we composite the rendered foreground with a fixed sky background for each trajectory. Specifically, we render each frame twice using black and white backgrounds, estimate the per-pixel alpha from the two renderings, and then blend the foreground with a sky image in screen space. This strategy preserves the original 3DGS appearance while producing more natural aerial observations for horizontally oriented cameras. 

\subsubsection{Isaac Sim-Based Data Collection Platform.}
\begin{figure}[tb]
  \centering
  \includegraphics[width=\linewidth]{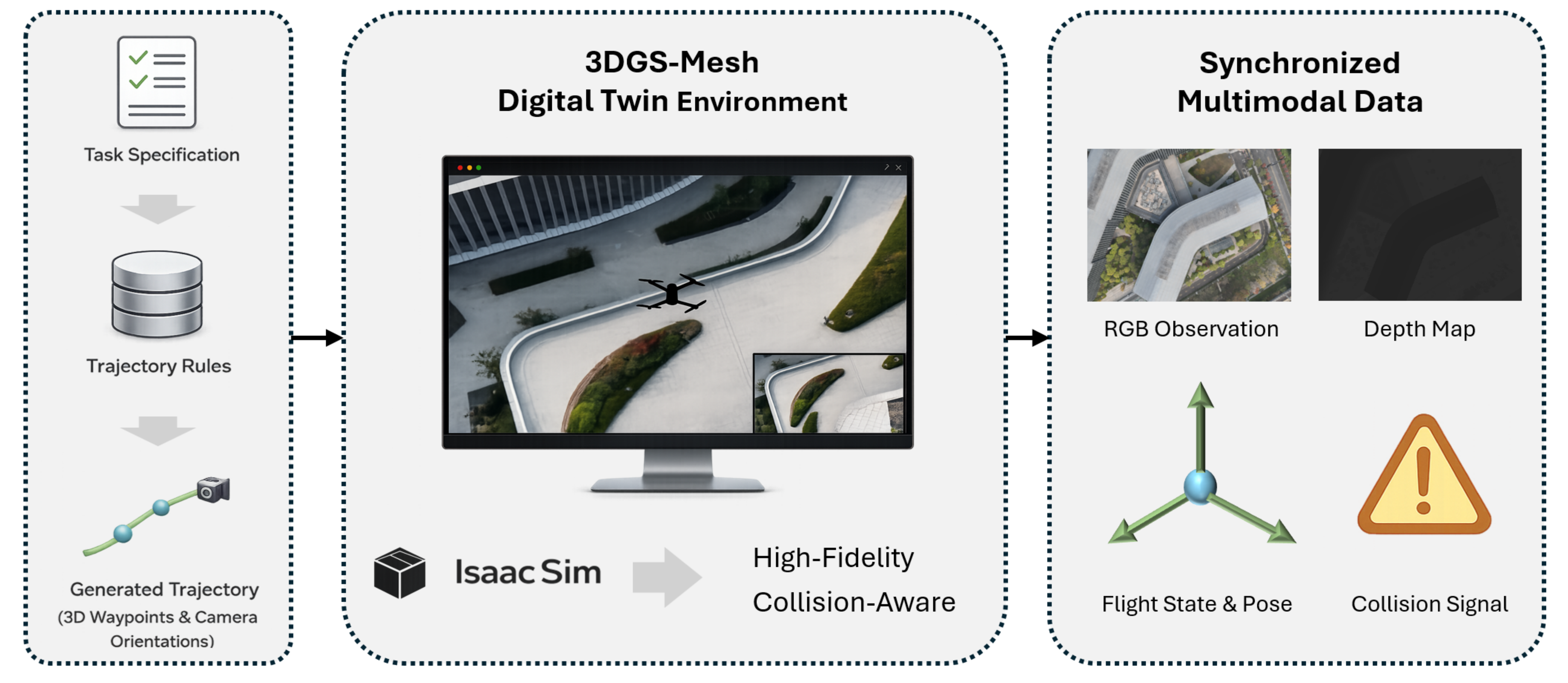}
\caption{\textbf{Isaac Sim-based data collection platform.} The platform generates trajectory data from a task specification and executes it in a 3DGS-Mesh digital twin environment in Isaac Sim. 
It collects synchronized multimodal data, including RGB observations, depth maps, flight states, poses, and collision signals.}
  \label{isaac}
\end{figure}

Figure~\ref{isaac} illustrates our Isaac Sim-based data collection platform~\cite{isaacsim}. Given a task specification, we first generate a reference trajectory consisting of 3D waypoints and associated camera orientations using our task-specific trajectory rules or the obstacle-aware planner described above. The aligned 3DGS--mesh digital twin is then imported into Isaac Sim, where the mesh serves as the collision-enabled physical scene and the 3DGS representation provides high-fidelity visual observations. During execution, the UAV follows the generated reference trajectory inside the simulator, and the platform records synchronized multimodal data streams at each step, including RGB observations, depth maps, ego poses, flight states, and collision signals.

This design offers three key advantages:
\begin{itemize}
    \item It preserves metric consistency between real-world acquisition, planning, and simulation, allowing all logged data to remain in a unified coordinate frame.
    \item It enables collision-aware evaluation through the mesh-based physical scene, which is not available in purely appearance-driven renderers.
    \item It supports scalable and programmable data generation: once a trajectory rule is defined, the simulator can automatically instantiate diverse initial states, execute the trajectory, and export the full multimodal sequence.
\end{itemize}

\begin{figure}[tb]
  \centering
  \includegraphics[width=\linewidth]{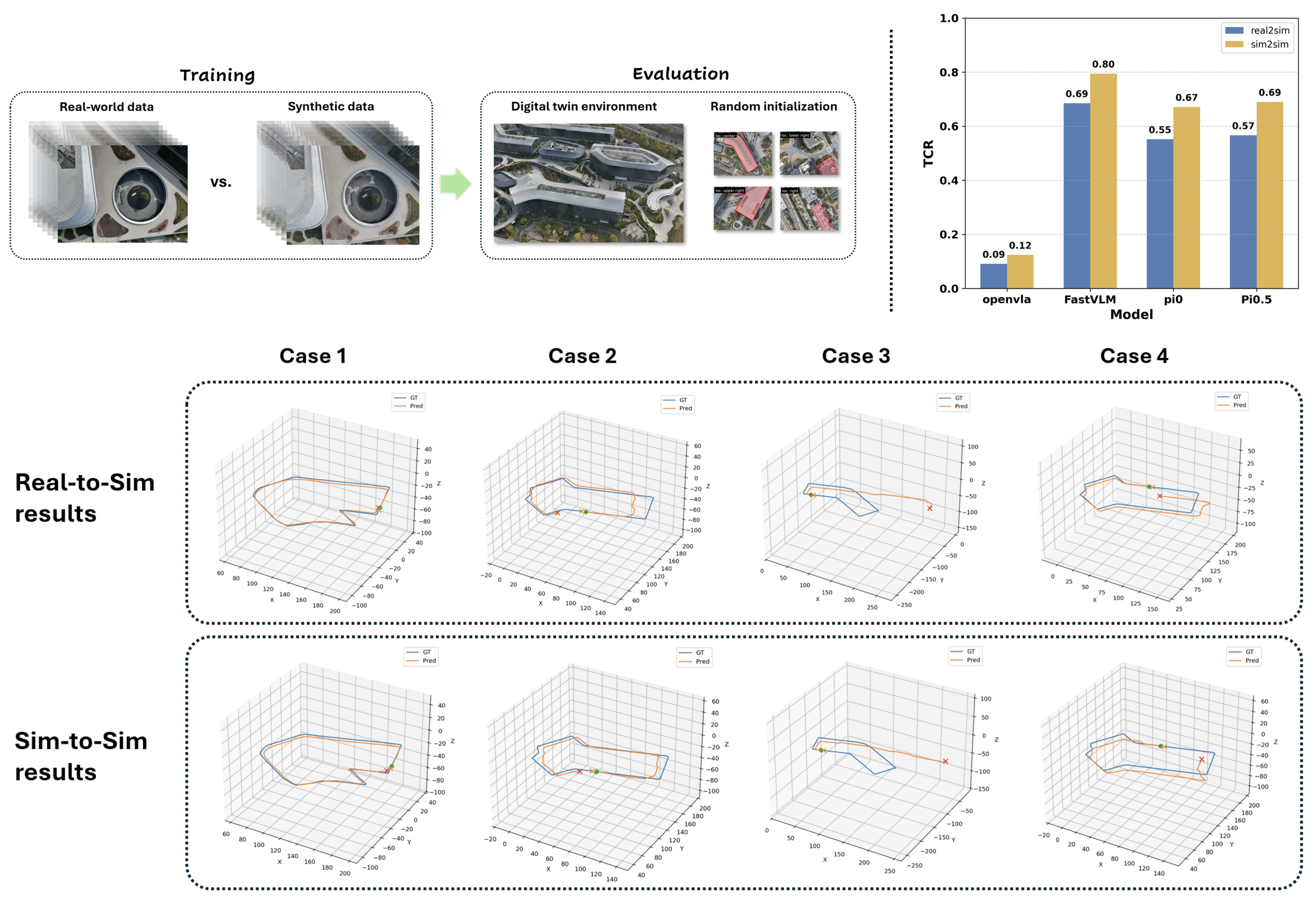}
\caption{\textbf{Overview and results of the real-to-sim and sim-to-sim comparison experiments.} Top-left: the experimental pipeline. Models are trained using either real-world data (real2sim) or synthetic data (sim2sim), and then evaluated in a digital twin environment with multiple random initializations. Top-right: quantitative test results showing the trajectory coverage rate (TCR) across different models under both settings. Bottom: qualitative trajectory comparisons for several representative test cases, with all results obtained from models trained on $\pi_0$. For each case, the predicted trajectory is compared with the ground truth in both real-to-sim and sim-to-sim evaluations. }
  \label{real}
\end{figure}

\subsubsection{Model and Training Details.}
We evaluate several representative VLA baselines on HUGE-Bench, including OpenVLA~\cite{kim2024openvla}, $\pi_0$~\cite{pi2025pi0}, $\pi_{0.5}$~\cite{pi2025pi05}, and FastVLM~\cite{vasu2025fastvlm}. Following prior work~\cite{gao2025openfly,wang2025uavflow}, we use both the first frame of the trajectory and the current observation as visual inputs, which help the model infer relative displacement and task progress in 3D space.

For $\pi_0$ and $\pi_{0.5}$, we set the action horizon to 20. During inference, each model predicts a 20-step action chunk, and we execute the averaged actions over the first 10 predicted steps to improve temporal stability. We initialize both models from their corresponding base checkpoints and finetune them for 5 epochs with a per-GPU batch size of 64. Unless otherwise specified, all remaining hyperparameters follow the default settings of the original $\pi_0$ implementation.

For FastVLM, we construct a VLA baseline by combining Apple's open-source FastVLM with the $\pi_0$ action expert (Gemma 300M). Specifically, the first-frame image, the current-frame image, and the user instruction are encoded by FastVLM, and the hidden states from its final layer are used as the conditioning context for the $\pi_0$ action expert, serving as the past keys and values in the KV cache. The robot state and noisy actions are concatenated and fed into the action expert as queries. We train the model with flow matching and perform 10 denoising steps at inference time to generate the final action sequence. The action horizon is set to 20. We finetune the model for 5 epochs with a per-GPU batch size of 32 and a learning rate of $2.5\times10^{-5}$. Optimization is performed with AdamW using betas $(0.9, 0.95)$ and weight decay $1\times10^{-10}$, together with a cosine learning rate scheduler and 1,000 warmup steps. We do not use LoRA and instead finetune all parameters end-to-end.

For OpenVLA, we use a per-GPU batch size of 18 and a learning rate of $5\times10^{-4}$. We enable LoRA finetuning with rank 32, while keeping all other training configurations the same as in prior work~\cite{wang2025uavflow,kim2024openvla}.

For the memory-augmented baseline, we evaluate MemoryVLA~\cite{shi2025memoryvla} to study whether long-horizon history helps high-level UAV VLA execution. 
Following the same training and evaluation protocol as OpenVLA, MemoryVLA takes the language instruction and visual observations as input, while additionally incorporating historical context through an implicit memory mechanism. 
We also test explicit frame-history grids sampled from the past 20 or 50 meters, which perform worse, due to shifted visual input distribution and temporal misalignment. This suggests that effective memory integration remains non-trivial, and our benchmark can support future memory-augmented VLA research.

For the depth-aware baseline, we build on $\pi_{0.5}$ and augment the visual observation with depth maps rendered from the aligned 3DGS--Mesh digital twin. 
The depth input is temporally aligned with the RGB observation and is used as an additional visual modality for action prediction. 
All other training settings follow the original $\pi_{0.5}$ baseline.

\subsubsection{Expert-Pilot Performance.}
To provide a human upper bound, we ask professional UAV pilots to execute each task according to the instruction and the initial UAV pose from the test set. 
For each episode, the pilot observes the task instruction and the initial state, and then performs the corresponding UAV operation. 
We then compute the same evaluation metrics between the human-executed trajectory and the ground-truth trajectory. 
This expert-pilot evaluation provides a reference for task feasibility and helps distinguish model limitations from ambiguity in the high-level instructions.

\section{Additional Analysis}
\label{sec:detailed_experimental_results}
\subsubsection{Real-to-Sim Analysis.}
We further study the gap between real-world and synthetic training data. Specifically, we collect 500 real UAV trajectories in the physical world for Inspection-B task and render their synthetic counterparts in the aligned 3DGS environment using the same camera intrinsics and extrinsics. All models are then evaluated in the same simulation platform for a controlled comparison, as illustrated in Fig.~\ref{real}. The quantitative results show two main observations. First, the gap between models trained on real data and those trained on synthetic data is relatively small, while models trained on synthetic data often achieve slightly higher scores under simulation evaluation. We attribute this to the higher distribution consistency between the synthetic training data and the simulation-based test environment, as well as the absence of real-world sensor noise and appearance variations. Second, the performance trends are largely consistent between real-data training and synthetic-data training, suggesting that training on our synthetic data can reflect the relative effectiveness of models trained on real trajectories. This indicates that HUGE-Bench can serve as a practical and scalable proxy for studying model behavior before moving to more expensive real-world collection.

\begin{table}[tb]
\centering
\small
\setlength{\tabcolsep}{5pt}
\renewcommand{\arraystretch}{1.15}
\caption{\textbf{Results on different instruction levels.}
We compare high-level and low-level instruction settings across different models on HUGE-Bench. 
Avg. TCR, nDTW, NTP, CR, and CSPL are reported.}
\label{tab:instruction_level_results}
\resizebox{\linewidth}{!}{%
\begin{tabular}{l|c|ccccc}
\toprule
\multirow[c]{2}{*}{\textbf{Methods}}
& \multirow[c]{2}{*}{\textbf{Instruction Level}}
& \multicolumn{5}{c}{\textbf{HUGE-Bench VLA}} \\
\cmidrule(lr){3-7}
& & Avg. TCR $\uparrow$ & nDTW $\uparrow$ & NTP $\uparrow$ & CR $\downarrow$ & CSPL $\uparrow$ \\
\midrule
\multirow[c]{2}{*}{OpenVLA}
& Low-level  & 0.147 & 0.013 & 0.167 & 0.008 & 0.119 \\
& High-level & 0.112 & 0.011 & 0.122 & 0.003 & 0.032 \\

\multirow[c]{2}{*}{FastVLM}
& Low-level  & 0.476 & 0.355 & 0.761 & 0.096 & 0.478 \\
& High-level & 0.285 & 0.136 & 0.633 & 0.060 & 0.391 \\

\multirow[c]{2}{*}{$\pi_0$}
& Low-level  & 0.613 & 0.561 & 0.752 & 0.008 & 0.889 \\
& High-level & 0.558 & 0.443 & 0.653 & 0.011 & 0.805 \\

\multirow[c]{2}{*}{$\pi_{0.5}$}
& Low-level  & 0.698 & 0.582 & 0.781 & 0.009 & 0.910 \\
& High-level & 0.581 & 0.467 & 0.618 & 0.018 & 0.805 \\
\bottomrule
\end{tabular}%
}
\end{table}

\begin{figure}[tb]
\centering
\includegraphics[width=0.75\linewidth]{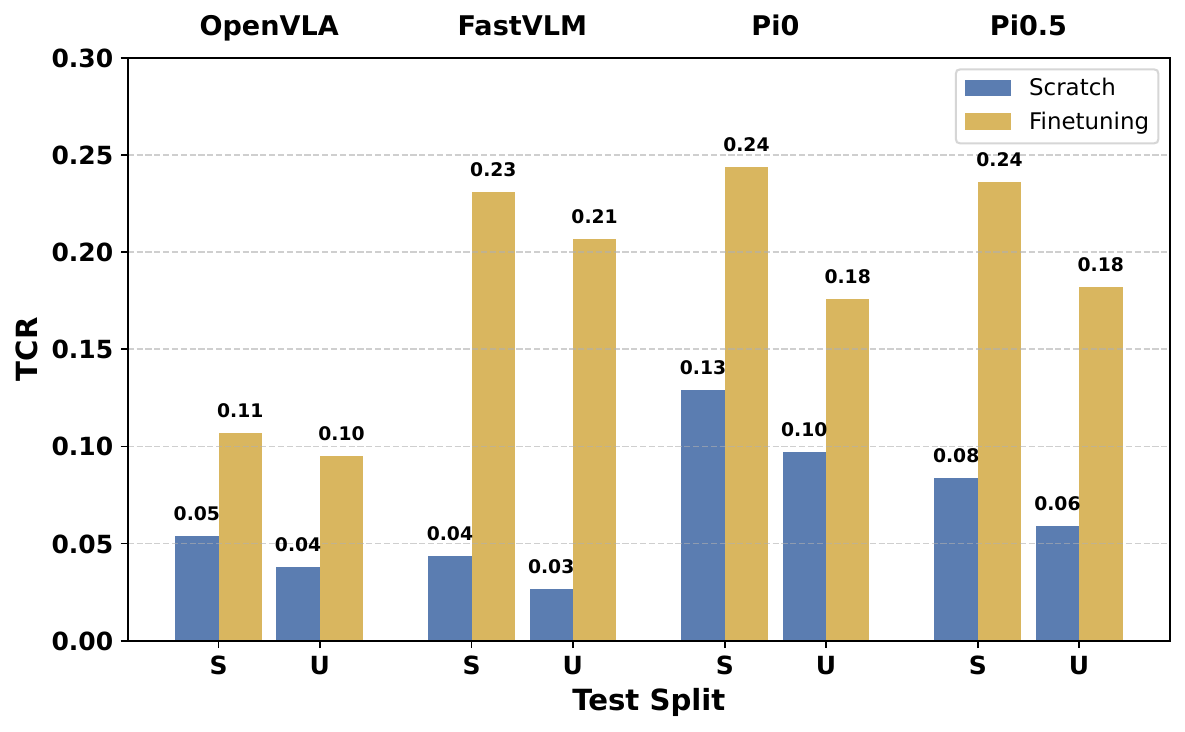}
\caption{\textbf{Comparison across different models on two test splits.} Blue bars denote training from scratch, while yellow bars denote finetuning.}
\label{scratch}
\end{figure}

\subsubsection{High-Level vs. Low-Level Instructions.}
Table~\ref{tab:instruction_level_results} compares models trained with traditional low-level instructions and our proposed high-level instructions across different models on HUGE-Bench (see Fig.~\ref{fig3}). We can see that low-level instructions generally lead to better performance, since they provide more explicit step-by-step guidance and reduce the ambiguity of action execution. In contrast, high-level instructions require the model to infer missing intermediate steps, maintain task progress over time, and decompose abstract goals into executable subtasks. The performance gap therefore highlights the additional difficulty introduced by high-level UAV VLA tasks, and further motivates future research on hierarchical reasoning, subtask decomposition, and long-horizon planning under compact natural language commands.

\subsubsection{Finetuning vs. Training from Scratch.}
Figure~\ref{scratch} compares models initialized from pretrained checkpoints with those trained from scratch on Inspection-B task. We observe that finetuning consistently yields better performance across models and splits. This result suggests that the knowledge acquired from large-scale embodied and robotic data is transferable to aerial UAV scenarios, despite the domain gap in viewpoint, motion pattern, and task formulation. These results indicate that leveraging pretrained VLA or robotic foundation models is a more effective strategy than training high-level UAV policies entirely from scratch.

\begin{figure}[tb]
  \centering
  \includegraphics[width=\linewidth]{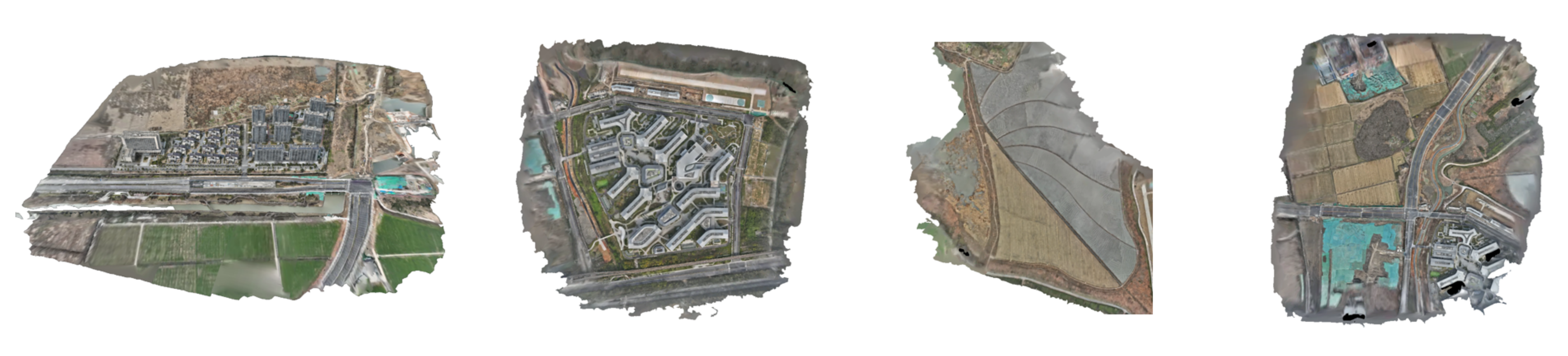}
\caption{\textbf{More visualizations of the 3D scenes in HUGE-Bench.} }
  \label{more_vis}
\end{figure}

\begin{figure}[tb]
  \centering
  \includegraphics[width=\linewidth]{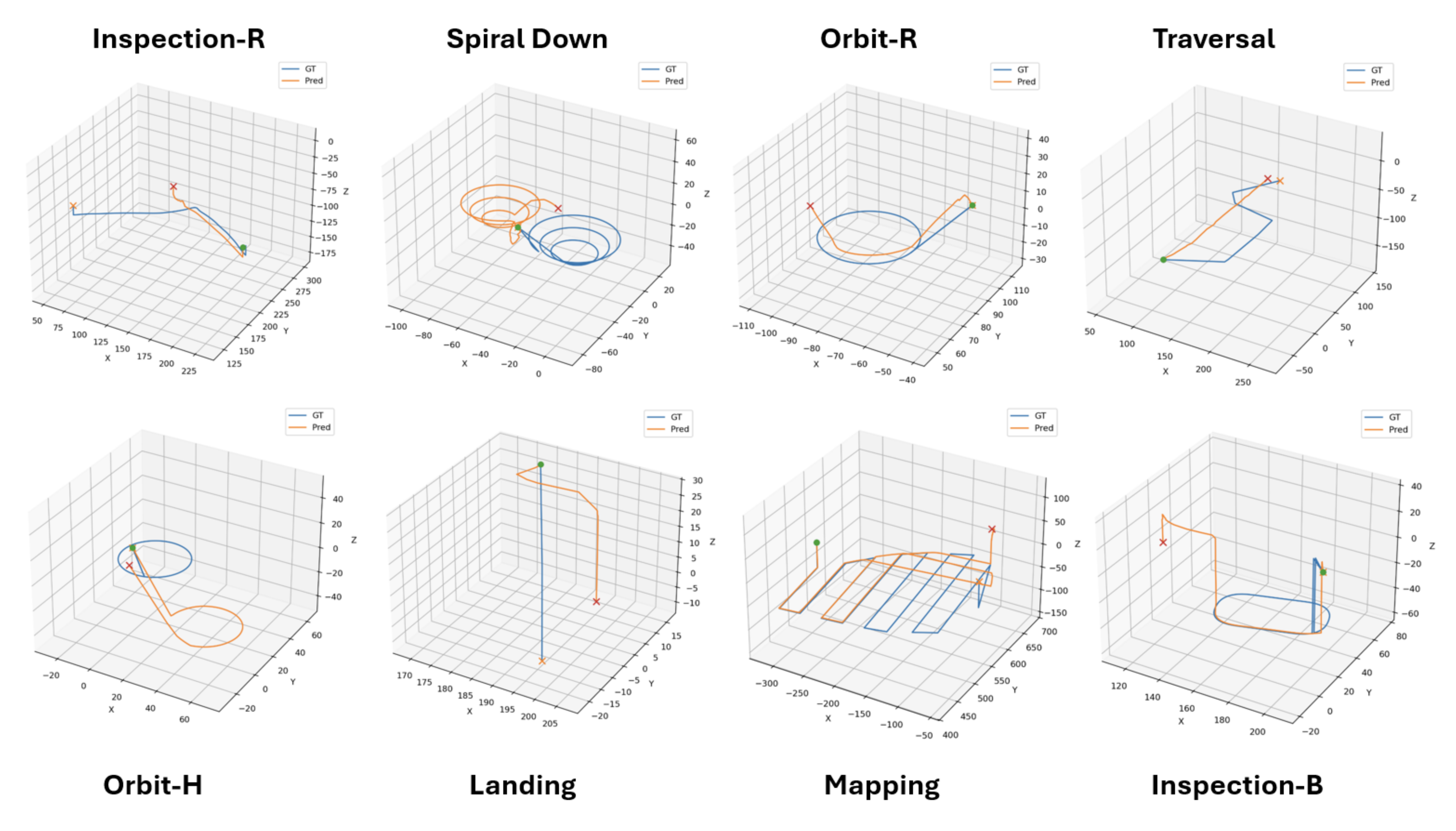}
\caption{\textbf{Visualization of failure results from $\pi_0$ across different tasks.} }
  \label{fail}
\end{figure}

\section{More Visualizations}
We provide additional visualizations of the reconstructed scenes in Fig.~\ref{more_vis}, illustrating the diversity of environments included in HUGE-Bench. 
We further show representative failure cases of $\pi_0$ across different task categories in Fig.~\ref{fail}. 
These examples reveal three major failure modes:

1) the model may localize the wrong target in landmark-centric tasks such as Orbit, Spiral Down, and Landing, leading to trajectories that are geometrically plausible but semantically incorrect. 

2) in long-horizon tasks such as Mapping and Inspection, the model may lose track of task progress, resulting in incomplete coverage, premature termination, or repeated local behaviors. 

3) in Traversal, the model may produce unsafe 3D execution and collide with obstacles, even when it roughly follows the intended direction.

These failures suggest that current VLA models still require stronger aerial-view visual grounding, progress-aware memory, and 3D spatial reasoning for reliable high-level UAV execution. 
More visualizations of the trajectory data and the Isaac Sim platform are provided on our project page.

\section{Scalability}
HUGE-Bench is scalable in four aspects. First, it supports scene diversity, and users can also build the benchmark on their own scenes. Second, it supports task diversity, as new high-level UAV tasks can be introduced by defining new task rules. Third, it supports trajectory diversity by varying the initial position, relative target direction, altitude, path smoothness, flight speed, camera intrinsics, and observation views. Finally, it supports data collection diversity, where trajectories can be obtained either from programmable rule-based generation or from human teleoperation.

\clearpage

\bibliographystyle{splncs04}
\bibliography{main}

\end{document}